\documentclass[conference]{IEEEtran}
\IEEEoverridecommandlockouts
\usepackage{cite}
\usepackage{amsmath,amssymb,amsfonts}
\usepackage{algorithmic}
\usepackage{graphicx}
\usepackage{textcomp}
\usepackage{xcolor}
\usepackage{multirow}
\def\BibTeX{{\rm B\kern-.05em{\sc i\kern-.025em b}\kern-.08em
    T\kern-.1667em\lower.7ex\hbox{E}\kern-.125emX}}
\begin{document}

\title{LAFA: Agentic LLM-Driven Federated Analytics over Decentralized Data Sources}

\author{
\IEEEauthorblockN{Haichao Ji}
\IEEEauthorblockA{
\textit{Shanghai Jiao Tong University}\\
Shanghai, China \\
haichao.ji2000@sjtu.edu.cn}
\and
\IEEEauthorblockN{Zibo Wang}
\IEEEauthorblockA{\textit{Peng Cheng Laboratory}\\
Shenzhen, China \\
wangzb@pcl.ac.cn}
\and
\IEEEauthorblockN{Cheng Pan}
\IEEEauthorblockA{
\textit{The University of Hong Kong}\\
Hong Kong, China \\
cpanpan@connect.hku.hk}
\and
\IEEEauthorblockN{Meng Han}
\IEEEauthorblockA{
\textit{Zhejiang University}\\
Hangzhou, Zhejiang, China \\
menghan@zju.edu.cn}
\and
\IEEEauthorblockN{Yifei Zhu}
\IEEEauthorblockA{
\textit{Shanghai Jiao Tong University}\\
Shanghai, China \\
yifei.zhu@sjtu.edu.cn}
\and
\IEEEauthorblockN{Dan Wang}
\IEEEauthorblockA{
\textit{The Hong Kong Polytechnic University}\\
Hong Kong, China \\
csdwang@comp.polyu.edu.hk}
\and
\IEEEauthorblockN{Zhu Han}
\IEEEauthorblockA{
\textit{University of Houston}\\
Houston, TX, USA \\
zhan2@uh.edu}
\thanks{This work is supported by the National Key R\&D Program of China (Grant No. 2023YFB2704400). Corresponding author: Yifei Zhu.}
}

\maketitle

\begin{abstract}
Large Language Models (LLMs) have shown great promise in automating data analytics tasks by interpreting natural language queries and generating multi-operation execution plans. 
However, existing LLM-agent-based analytics frameworks operate under the assumption of centralized data access, offering little to no privacy protection. 
In contrast, federated analytics (FA) enables privacy-preserving computation across distributed data sources, but lacks support for natural language input and requires structured, machine-readable queries.
In this work, we present LAFA, the first system that integrates LLM-agent-based data analytics with FA. LAFA introduces a hierarchical multi-agent architecture that accepts natural language queries and transforms them into optimized, executable FA workflows. A coarse-grained planner first decomposes complex queries into sub-queries, while a fine-grained planner maps each sub-query into a Directed Acyclic Graph of FA operations using prior structural knowledge. To improve execution efficiency, an optimizer agent rewrites and merges multiple DAGs, eliminating redundant operations and minimizing computational and communicational overhead.
Our experiments demonstrate that LAFA consistently outperforms baseline prompting strategies by achieving higher execution plan success rates and reducing resource-intensive FA operations by a substantial margin. This work establishes a practical foundation for privacy-preserving, LLM-driven analytics that supports natural language input in the FA setting.
\end{abstract}

\begin{IEEEkeywords}
LLM agents, federated analytics, privacy computing
\end{IEEEkeywords}

\section{Introduction}

The rapid development of Large Language Models (LLMs) has offered unprecedented capabilities in natural language understanding, reasoning, and planning \cite{yao2023react}, significantly transforming the landscape of data analytics. 
LLMs can interpret complex analytical intents, generate structured code, and orchestrate multi-step tasks by interacting with external environments such as databases and computation sandboxes.
These capabilities have led to the emergence of LLM-based agents that decompose high-level queries, plan analytical workflows, and execute or verify results through tool interactions. 
Notably, recent agent systems such as AutoTQA \cite{zhu2024autotqa} have demonstrated the feasibility of end-to-end data analytics utilizing LLMs, where queriers only need to ask in natural language and LLMs can transform questions into executable operations.

In recent years, regulations such as the General Data Protection Regulation (GDPR) in the EU \cite{EuropeanParliament2016a} and the California Consumer Privacy Act (CCPA) in California \cite{CAdataregulations} have introduced strict constraints on personal data access and processing, reshaping the requirements for data analytics workflows.
However, typical LLM-based systems rely on centralized architectures where data is aggregated at servers or exposed via external APIs, posing substantial privacy risks. 
These risks include unintended memorization, misuse during transmission or storage, and a lack of transparency.
Consequently, current LLM-agent analytics frameworks that assume unrestricted access to raw data are ill-suited for real-world deployments where privacy is a fundamental requirement.


Federated analytics (FA) \cite{wang2025survey} has emerged as a promising paradigm that enables data analytics across distributed user devices without collecting raw data into a central repository. 
FA frameworks enable each client to compute locally over private data and share only privacy-preserving intermediate results with a central server, which then synthesizes the global answer. It can be further augmented with a range of privacy-preserving techniques, such as differential privacy (DP) \cite{dwork2006calibrating}, secure multi-party computation (SMPC) \cite{ben2008fairplaymp}, and homomorphic encryption (HE) \cite{naehrig2011can}, to provide strong protection against information leakage. 
However, prior work in FA has primarily concentrated on algorithmic designs tailored to specific data science tasks under privacy and communication constraints. In contrast, enabling support for complex, natural language queries remains largely unexplored, posing a significant barrier to practical adoption in real-world settings.


To bridge this gap, we propose integrating FA with LLM-agent-based data analysis, enabling complex query resolution in a privacy-preserving and natural language-supported manner. While this approach holds substantial promise, our in-depth analysis reveals that LLM-driven federated analytics still faces two fundamental challenges.
First, the analytics workflow generated by LLM agents often contains logical deficiencies, including disrupting orders and missing operations.
Current LLM agents, even those based on advanced prompting techniques like ReAct \cite{yao2023react}, struggle to reason about the specialized workflows of FA.  
In practice, we observe that agents may generate incomplete, semantically incorrect, or even computationally infeasible plans. For example, they may attempt to apply preprocessing after encryption, or add noise before decryption, all of which are invalid operations in the FA protocol. 
Second, existing LLM agent solutions still have limited capability in efficiently decomposing practical queries, which are usually complex and contain multiple sub-intents.
They tend to replicate redundant operations (e.g., repeatedly encrypting the same data for each sub-query), wasting reusable intermediate results.
These issues not only degrade the correctness of the final answer but also incur substantial computational and communication overhead, particularly problematic in large-scale federated environments.

To address these challenges, in this work, we propose LAFA, an LLM-driven federated analytics framework for natural-language-enabled privacy-preserving data analytics over decentralized data.  
LAFA introduces a hierarchical multi-agent framework that transforms natural language queries into executable and optimized FA operations. 
It first incorporates a coarse-grained planner to segment complex queries into single analytical sub-queries. A fine-grained planner, equipped with a repository of FA DAG templates, then maps each sub-query to a preliminary FA DAG using stored structural priors, ensuring each decomposition operation adheres to valid FA semantics.  An optimizer agent is then designed to rewrite and merge multiple preliminary DAGs into an optimized one by eliminating redundant operations. 

Overall, our main contribution is summarized as follows: 

\begin{itemize}
    \item LAFA is the first LLM-driven federated analytics framework that enables efficient complex query processing in natural languages with privacy preserved. 
    \item We design a hierarchical agent system that transforms natural language queries into executable federated analytics operations with reduced sequencing errors. 
    \item We design a DAG optimizer tailored for LLM-generated FA workflows with significantly reduced computational and communication overhead. 
    \item Extensive experiments on real-world complex queries demonstrate that LAFA's superior performance in query decomposition quality, privacy compliance, and resource efficiency.
\end{itemize}

\section{Background}
\subsection{Large language models for data analytics}
The advent of LLMs has significantly transformed natural language understanding, generation, and interactive decision-making. LLMs demonstrate remarkable capabilities in reasoning, planning, memorization, and tool use \cite{yao2023react}.
Their success across diverse domains has motivated researchers to explore their application beyond language tasks, such as in data analytics. 
However, while LLMs exhibit impressive reasoning abilities, leveraging them for systematic and verifiable data analytics presents unique challenges, such as hallucination and limited numerical accuracy.

\textbf{LLM prompting for analytical reasoning}
Prompt engineering has emerged as a fundamental technique to guide LLMs toward desired behaviors. 
Techniques like Chain-of-Thought (CoT) \cite{wei2022chain} prompting encourage models to explicitly reason through intermediate operations, improving performance on complex reasoning tasks.
Reasoning and Acting \cite{yao2023react} prompting extends this idea by interleaving reasoning traces with actions that query external environments, such as databases or APIs. 
This synergy between internal deliberation and external interaction has been shown to improve factual correctness, robustness, and transparency in tasks like question answering and decision making.
Building on these prompting paradigms, data analytics systems increasingly adopt structured prompting strategies that encourage LLMs to decompose high-level analytical queries into sequential subtasks involving both reasoning and tool use. 

\textbf{LLM agents for complex data task decomposition} 
Moving beyond single-operation prompting, researchers have developed LLM-based agents that autonomously plan, reason, act, and reflect across multiple operations to solve complex tasks. 
In the agent paradigm, an LLM (or multiple LLMs) operates as a decision-making entity that dynamically generates subgoals, executes actions, observes outcomes, and iteratively refines its strategy. 
Agents collectively transform vague user queries into precise execution plans, operate external tools (e.g., SQL engines, Python sandboxes), and verify intermediate results.


\subsection{Privacy-preserving data analytics}
\textbf{Federated analytics.} FA is a collaborative data analytics framework that supports secure and private data analytics across multiple decentralized clients without collecting raw data into a central repository. 
In traditional FA settings, a server initiates a structured process to answer a specific analytical query, typically in structured forms such as count-mean sketches \cite{Apple_DP_2017} and hypothesis tests \cite{liu2024measuring}.

The workflow begins when the server transforms the analytical query into client-level machine-executable tasks for all clients. 
Rather than moving data to the central server for task processing, each client processes the tasks locally over its private data. 
This local computation produces intermediate results that encapsulate essential information needed to answer the query while preventing the disclosure of raw data.
Clients then securely transmit these locally computed results to the server, where a specialized aggregator combines them into a final result. 
Through careful aggregation techniques, the server can synthesize a collective analytical answer that accurately reflects the global data statistics without direct access to any client’s data, which can then be released to the querying party.


\textbf{Privacy protection techniques in FA.}
To strengthen privacy guarantees in FA, a variety of cryptographic and algorithmic techniques have been developed and integrated. 
The most prominent approaches include DP, HE, and SMPC. DP adds carefully calibrated random noise to the outputs of computations. 
This ensures that the inclusion, removal, or alteration of a single individual’s data point has a minimal impact on the final analytical result, thereby masking any single participant’s contribution.
HE allows computations to be performed directly on encrypted data without requiring decryption.
For instance, in additive homomorphic encryption (AHE), the sum of two plaintext values can be obtained by applying a special operation to their ciphertexts: $enc(a)\bigoplus enc(b) = enc(a+b)$ where $enc()$ represents the encryption function and $\bigoplus$ denotes the ciphertext-level addition. 
SMPC enables multiple clients to jointly compute a function over their private inputs while revealing no additional information beyond the final result. 

\section{System design}

\subsection{Design goals and challenges}
LAFA aims at bridging the gap between natural language data analytics and privacy-preserving computation to support natural-language-supported privacy-preserving decentralized data analytics. 
Essentially, it acts as an LLM-powered compiler for FA, transforming complex natural language queries into optimized, privacy-preserving execution pipelines.
The system is carefully designed to address the following key challenges:
\begin{itemize}
    \item Logical sequencing deficiency: LLM agents often generate FA operation sequences that violate required procedural semantics, leading to invalid or insecure computations. 
    \item  Multi-sub-query decomposition: complex natural language queries frequently contain multiple analytical intents that must be decomposed into coordinated FA pipelines.
    Without structural guidance, LLM agents tend to generate redundant FA operations across sub-queries.
\end{itemize}


\subsection{Design overview}
\begin{figure}[t]

\centering

\includegraphics[width=0.9\linewidth]{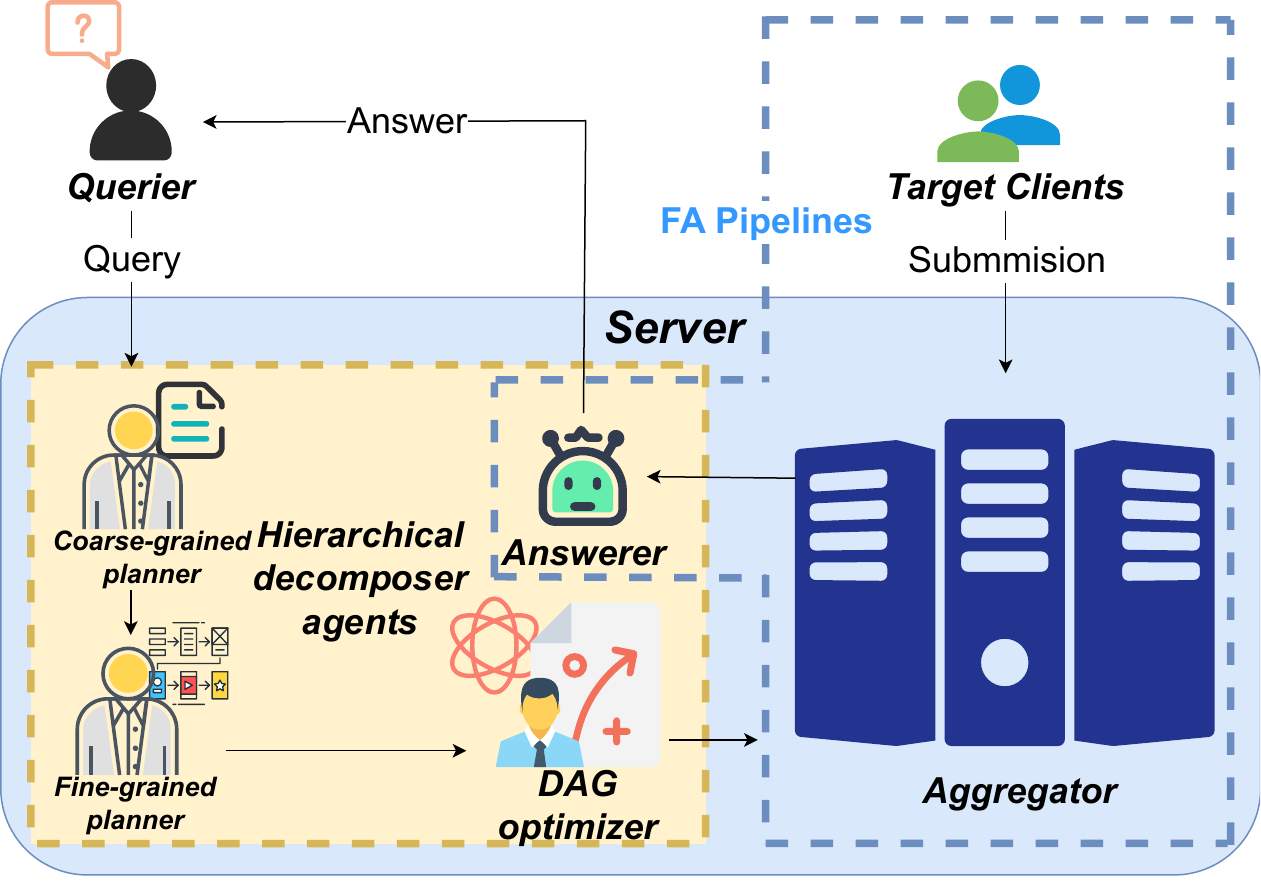}
\caption{The system overview of LAFA}
\label{figSystemOverview}
\end{figure}

To address these challenges, LAFA follows a modular design philosophy, consisting of a hierarchical multi-agent framework layered on top of an FA engine.
LAFA is structured around three main entities: the queriers, the target clients, and the server, and three core stages: hierarchical decomposition, DAG optimization, and FA execution, as illustrated in Fig. \ref{figSystemOverview}.  Each entity plays a specialized role in the FA process:

\textbf{Queriers.} They are entities such as researchers, companies, or system administrators interested in exploring aspects of client data through structured queries.
They issue natural language queries to the server, which returns corresponding answers also in natural language.

\begin{figure*}[t]

\centering

\includegraphics[width=1.0\linewidth]{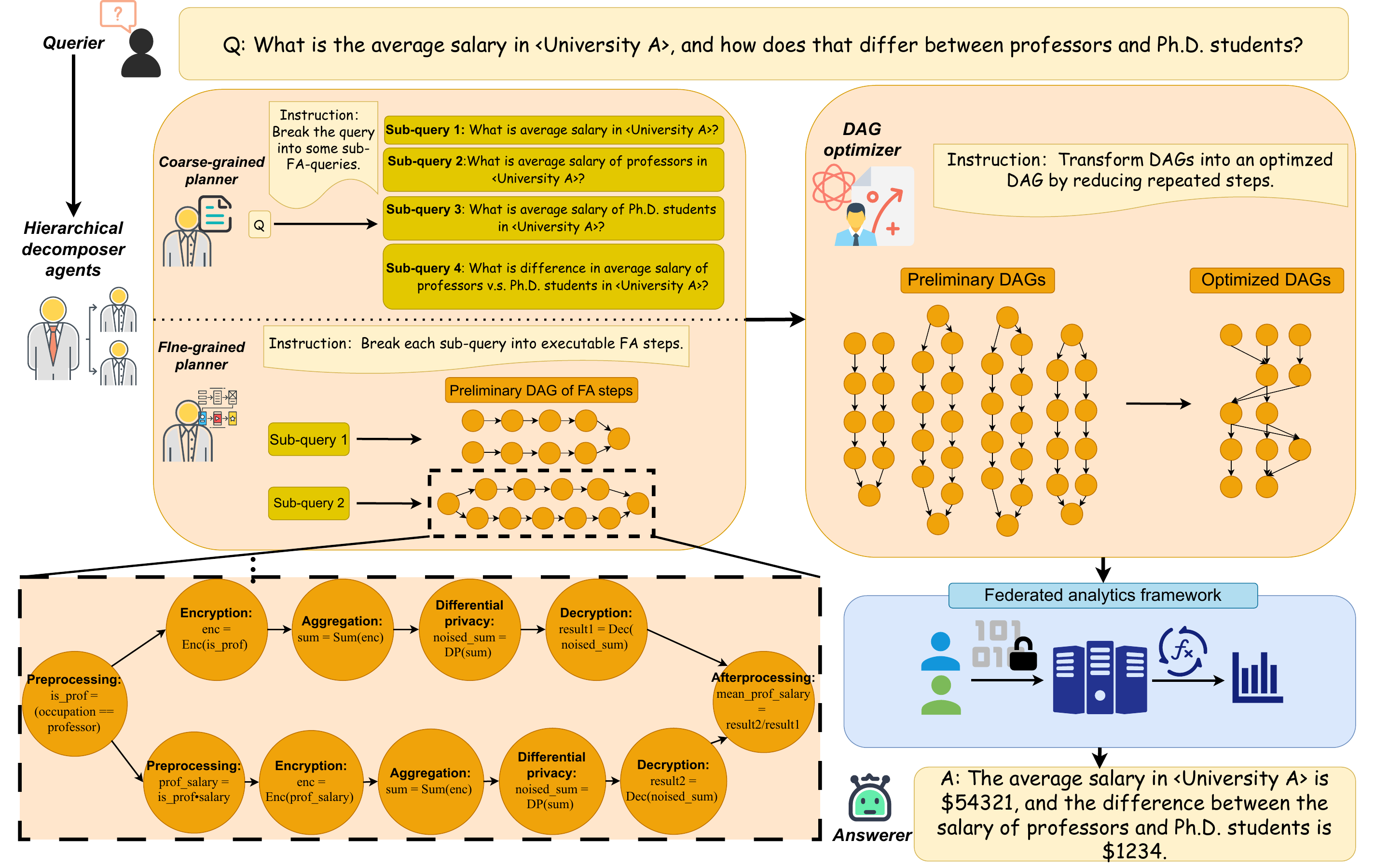}
\caption{The workflow of LAFA using a query as an example}
\label{figExecution}

\end{figure*}
\textbf{Clients with devices.} A collection of distributed users with smart devices. 
These devices participate in FA tasks by contributing locally processed and encrypted data. 
Each client device in LAFA is equipped with the capability to perform local data filtering, secure encryption, and lightweight computation.
They range from powerful computing platforms such as laptops to more resource-constrained devices such as smartphones.
These modern smart devices are increasingly equipped with sufficient computation power, well-suited for local processing tasks required by LAFA.
This ensures that sensitive raw data remains strictly local, aligning with the privacy-by-design principle of FA.

In LAFA, the dataset is horizontally partitioned such that each client device holds exactly one data record. 
This design choice aligns with common deployment scenarios for FA, where data is inherently distributed across user-controlled devices or accounts.
Each record follows a consistent schema, containing a fixed set of features known to the system. This assumption of uniformity simplifies query planning and ensures that the same FA workflow can be applied consistently across the client population.

\textbf{Server.} The server in LAFA is a coordination node that hosts an aggregator and a multi-agent system. 
The aggregator performs aggregation of encrypted data, applies DP noise, conducts decryption, and computes the final analytics result as designed in a typical FA pipeline. 
In parallel, since we observe that single-agent prompting often fails to decompose complex federated queries correctly and produces inefficient execution plans, we design a hierarchical multi-agent system that handles query decomposition, optimization, and execution planning. This hierarchical approach addresses the known weaknesses of LLM agents in handling complex multi-intent queries and enforcing correct FA operation sequencing.

\subsection{Hierarchical multi-agents design} 
We now introduce the detailed design of our three core agents, each guided by specifically crafted prompts to fulfill distinct roles in transforming natural language queries into optimized executable FA workflows.

\textbf{Coarse-grained planner agent.} It is responsible for identifying and segmenting a given complex query into multiple single-query components, ensuring that each sub-query can be handled separately.
We deliberately design its prompt to decompose the received natural language query into sub-queries corresponding to a single analytical intent, such as computing an average or comparing subgroups.
The coarse-grained planner agent is encouraged to recognize temporal, logical, or comparative structures in natural language (e.g., ``and'', ``vs.'') to produce sub-queries.

\textbf{Fine-grained planner agent.} Once the single queries are identified, it then further decomposes them into an executable FA operations DAG.
Its prompt design references the prior DAGs that encapsulate predefined FA pipelines, including operations and their sequences for various query types.
FA pipelines are summarized based on previous work \cite{margolin-2023-arboretum} as: preprocessing, encryption, aggregation, noise addition, decryption, and postprocessing.
In addition, it is instructed to select the correct DAG skeleton and generate preliminary DAGs by instantiating it with features of the data according to the sub-query.

\textbf{DAG optimizer agent.} After receiving all preliminary DAGs of each sub-query, an optimizer agent analyzes them and generates a unified posterior DAG that removes redundant operations (e.g., repeated encryption or aggregation). 
Its prompt is carefully designed to perform the following key functions:
\begin{itemize}
    \item Merging overlapping or related operations: It identifies and unifies operations that appear repeatedly across sub-queries, such as multiple encryptions of the same feature or repeated aggregations over similar subsets of data.
    \item Simplifying redundant or broad operations: The agent applies role-based refinements to generalize or consolidate operations. For example, if multiple sub-queries filter on overlapping roles (e.g., ``graduate students'' and ``Ph.D. students''), it reorganizes them under a shared filtering condition where applicable.
    \item Augmenting with implied operations: When the final query intent suggests additional steps, such as computing differences or ratios, the optimizer is instructed to insert new nodes to complete the analytical goal.
    \item Ensuring structural validity: It strictly maintains the logical and privacy-preserving execution order of FA, guaranteeing that the resulting DAG respects all data dependencies and adheres to secure processing constraints.
\end{itemize}

\textbf{Answerer}: After the optimized DAG is executed by the FA pipeline, an answerer agent composes the final answer in natural language. 
It retrieves the results of the FA pipeline and formats them into a coherent response according to the input query. 
It ensures that queriers can interact with the system entirely through natural language, without needing to understand underlying FA concepts.



\subsection{Execution workflow}
To illustrate LAFA’s full workflow, consider the example shown in Fig. \ref{figExecution} of the paper. Queries are submitted in natural language by end-users, as in the example, a data analyst is curious about the average salary in a university and the difference between that of professors and Ph.D. students.
This complex query contains multiple sub-intents that need to be accurately decomposed and jointly answered without duplicating computations.
It is received by the server and passed first to the coarse-grained planner.
It performs the rough decomposition of the query into several sub-queries.
Each sub-query is processed by the fine-grained planner with FA operations DAG knowledge. 
It will identify the task type of each sub-query and decompose it into a preliminary DAG of corresponding FA operations, which includes preprocessing (data filtering), encryption, aggregation, noise addition, decryption, and postprocessing (result computation).

The output preliminary DAGs for each sub-query are then processed by the DAG optimizer to generate an optimized DAG.
Before this optimization, there were repeated operations in each preliminary DAG. 
For example, in the Fig. \ref{figExecution}, the salary information of both professors and Ph.D. students is repeatedly collected, encrypted, and aggregated.
Repeated execution of FA operations like encrypted aggregation and MPC decryption is extremely resource-intensive. 
Na\"ive FA execution may involve many rounds, each costing MBs of data transfer, minutes of client CPU load, and TBs of server load\cite{roth-2020-orchard}. 
Efficient query planning and operation fusion are thus critical to avoid this waste.
The DAG optimizer rewrites and unites these preliminary DAGs into an optimized DAG without repeated operations. 
It achieves this by identifying the same FA operations and partitioning the clients into groups based on similar features.
For example, in the Fig. \ref{figExecution}, the people in the university will be separated into professors, Ph.D. students, and others.
By collecting information from these partitioned parties only once and final calculations, our system achieves the same goal but with less resource usage.

\textbf{Implementation of FA Primitives.}
We clarify that LAFA does not propose new federated analytics protocols or cryptographic primitives. 
Instead, it builds directly on standardized FA backends such as Arboretum \cite{margolin-2023-arboretum} and Orchard \cite{roth-2020-orchard}. 
Concretely, the FA backend in our setting includes the canonical pipeline of local preprocessing, secure encryption or secure aggregation, differential privacy noise addition, decryption, and lightweight postprocessing. 
These operations are already well studied and provide formal differential privacy guarantees under the assumptions in prior work.

Our threat model follows standard FA deployments: (1) clients are honest-but-curious and only contribute their local data record; (2) the server is semi-honest but does not see raw data, only encrypted or noise-protected aggregates; (3) the final released result is protected by differential privacy.

With these assumptions, LAFA inherits the privacy and security guarantees of the backend FA framework. 
The novelty of LAFA lies exclusively in the orchestration layer—translating natural language into valid FA workflows and optimizing DAGs. 
The execution and privacy guarantees are provided by the underlying FA system, and LAFA remains compatible with future improvements in FA primitives.

\section{Evaluation}

In this section, we comprehensively evaluate the design of LAFA in supporting LLM-agent-driven FA. Our evaluation seeks to answer the following research questions (RQs):

\begin{itemize}
    \item \textbf{RQ1}: Can LAFA correctly understand FA procedures and decompose complex, multi-intent queries into executable FA operations in the correct order?
    \item \textbf{RQ2}: Can LAFA reduce redundant or repeated FA operations in generated query plans, improving overall resource efficiency?
\end{itemize}

\subsection{Experimental setup}
\label{subsec:ExperimentSetup}
\textbf{Query dataset generation.} Due to the lack of a real-world complex natural language query dataset, we use GPT-4o to automatically generate 20 natural language analytical queries based on the dataset schema provided in a commonly used FA dataset and a real-world data privacy report from Apple. Our prompt design for query generation is listed in the Appendix.
\begin{itemize}
    \item \textbf{AdultPii}: We use a structured tabular dataset commonly employed in FA experiments \cite{programmingDP}, consisting of 32,563 records and 18 features per individual. The dataset includes a mix of personally identifiable information (PII), demographic attributes, and economic indicators, which makes it well-suited for evaluating privacy-preserving data analytics frameworks. 
    \item \textbf{Apple}: 
    We further synthesize complex FA queries based on real-world use cases documented in Apple’s differential privacy report\cite{apple2017dp} using a retrieval-augmented generation (RAG) approach. Specifically, we use the official Apple technical report as the retrieval source and provide it as context to an LLM. We then design structured prompts that guide the model to generate natural language queries reflecting typical analytics demands across domains such as emoji usage, Safari diagnostics, QuickType suggestions, and health data tracking. These queries naturally reflect diverse analytical goals involving multiple attributes, filters, and temporal or demographic comparisons, which frequently arise in real-world applications.
\end{itemize}

For illustration, below are representative queries from each dataset (more example queries are provided in the Appendix):
AdultPii examples:
\textit{``What is the average salary of all individuals, and how does it differ between male and female individuals?"}
and \textit{``What is the percentage of individuals working more than 40 hours per week, and how does this percentage change across different education levels?"}
Apple examples:
\textit{``What are the most frequently used emoji in messages this month, and how does their usage differ between users under 30 and over 50?"}
and \textit{``Which health data types are most frequently edited by users under 40, and how does this compare to users over 60?"}

\textbf{Evaluation metrics.} To evaluate the agent’s performance, we adopt the following key metrics:
\begin{itemize}
    \item Completion ratio: The proportion of queries for which the LLM agents generate a complete and valid FA execution DAG, which means the nodes in the DAG are executable operations in the FA and the edges in the DAG represent the appropriate order in the FA.
    \item Operation count: Total number of various types of operations per client in each DAG. It includes data access operations, encryption operations, aggregation operations (divided by client population), noise addition operations, decryption operations, and calculation operations.
\end{itemize}

\textbf{Baseline methods.} We compare LAFA against two common single-agent prompting strategies:

\begin{itemize}
    \item Zero-shot prompting: directly prompting the LLM by integrating prompts of all agents in LAFA using proper logic and transitions.
    \item One-shot prompting: providing the LLM an extra manually crafted example with sample input and output, besides the same prompt in zero-shot prompting.
\end{itemize}

\textbf{Quality evaluator.}  We implement a lightweight structure checker to validate the correctness of the generated FA operations. For nodes in the generated DAG structure, the checker verifies whether each node corresponds to a valid FA operation. For edges, the checker examines whether the execution order implied by the edges respects the logical dependencies of FA workflows. The correctness of intermediate and final computation results is further manually inspected to ensure the eventual correctness. 

\textbf{LLM backend.} All experiments are conducted using GPT-4 via the OpenAI API. The temperature parameter is universally set to 0.

\subsection{Completion ratio (RQ1)}
We measure the success rate of LAFA in producing valid DAGs across all queries. 
Compared to 10$\sim$15\% for zero-shot and 60$\sim$75\% for one-shot prompting, LAFA consistently attains a near-perfect completion ratio of 95$\sim$100\%, as in Table \ref{tab:result}, especially on queries involving multiple sub-questions or complex aggregation logic.
This indicates its superior ability to decompose and structure tasks effectively.

\begin{table*}[h]
    \centering
    \caption{Comparison of completion ratio and average operation count over 20 queries for each dataset.}
    \begin{tabular}{ccccccccc}
    \hline
        \multirow{2}{*}{\textbf{Dataset\&Query}} & \multirow{2}{*}{\textbf{Method}} & \multirow{2}{*}{\textbf{Completion Ratio}} & \multicolumn{6}{c}{\textbf{Average Operation Count}}\\
        & & & Acce & Enc & Aggr & DP & Dec & Cal \\
        \hline
        \multirow{3}{*}{AdultPii} & Zero-shot prompting & 15\% & - & - & - & - & - & -\\
        & One-shot prompting & 75\% & 3.55 & 3.55 & 3.55 & 4.95 & 4.95 & 3.85\\
        & LAFA & 100\% & 2.20 & 2.20 & 2.20 & 3.70 & 3.70 & 5.50\\
    \hline
        \multirow{3}{*}{Apple} & Zero-shot prompting & 10\% & - & - & - & - & - & -\\
         & One-shot prompting & 60\% & 3.40 & 3.40 & 3.40 & 4.60 & 4.60 & 3.45\\
         & LAFA & 95\% & 2.00 & 2.00 & 2.00 & 3.45 & 3.45 & 5.25\\
        \hline
    \end{tabular}
    \label{tab:result}
\end{table*}
Zero-shot prompting often produces incomplete execution plans, especially failing to include critical operations in the final calculation phase.
For example, it may correctly perform local filtering, encryption, aggregation, noise addition, and final decryption, but omit the final calculation required to obtain averages or ratios.
One-shot prompting improves over zero-shot by successfully capturing most required operations.
However, its performance remains suboptimal on complex queries, as it struggles to identify the relationships and distinctions among the sub-queries.
In contrast, LAFA demonstrates robust handling of complex multi-sub-query structures and respects the logical order of FA operations.
\subsection{Operation count (RQ2)}
We evaluate the average number of operations in the generated DAG of each query. 
We observe six operations in FA and denote them as Acce (Access to data per client), Enc (Encryption per client), Aggr (Aggregation times/clients population), DP (Noise addition for DP guarantee), Dec (Decryption), and Cal (Calculation operation). A lower operation count in computationally and communicationally intensive operations implies a more optimized and concise execution plan. 
LAFA consistently produces DAGs with fewer operations in them, as in Table \ref{tab:result}, reflecting better decomposition logic and aggregation reuse.
Compared with one-shot prompting, LAFA gains an average reduction of 1.35$\sim$1.40 in Acce, Enc, and Aggr operations and 1.15$\sim$1.25 in DP and Dec operations.

We observe that LAFA consistently achieves a lower operation count compared to one-shot prompting in resource-intensive operations. 
This is because LAFA avoids repeated access to user data, redundant encryption, and duplicate aggregation operations when multiple sub-queries focus on overlapping data. 
Instead of performing separate secure computations for each sub-query, LAFA reuses shared intermediate results and shifts the complexity to the final stage by applying more lightweight final calculations (e.g., division, subtraction, ratio computation).

\subsection{Ablation study}
We conduct ablation experiments on all datasets and queries to assess the contribution of each module in LAFA. We consider the following variants:
\begin{itemize}
    \item w/o preliminary DAG knowledge: removing feeding the planner preliminary DAGs of FA.
    \item w/o hierarchical planner: using one flat planner.
    \item w/o DAG optimizer: remove the DAG optimizer.
\end{itemize}
\begin{table}[h]
    \centering
    \caption{Completion ratio comparison.}
    \begin{tabular}{c|c}
    \hline
        Method & Completion ratio\\
        \hline
        w/o preliminary DAG knowledge & 0\\
        w/o hierarchical planner & 35\%\\
        w/o DAG optimizer & 100\%\\
        LAFA & 100\%\\
    \hline
    \end{tabular}
    \label{tab:completionRatioAbl}
\end{table}
As shown in Table \ref{tab:completionRatioAbl}, removing the preliminary DAG knowledge results in a 0\% completion ratio, indicating that the LLM agent fails to generate valid execution plans without access to structural priors.
All edge dependencies are misaligned, especially in the transitions between data filtering and encryption, as well as between decryption and subsequent computation.
This highlights the importance of grounding the agent in pre-defined FA procedural knowledge to enable even minimal task decomposition.
Disabling the hierarchical planner significantly degrades performance, reducing the completion ratio to 35\%. 
This suggests that a flat or non-modular planning approach cannot effectively handle multi-sub-query decomposition, leading to fragmented or incomplete DAGs.
It is worth mentioning that the robustness of the DAG optimizer contributes to the success of samples despite the weak performance of the flat planner, to recover part of the execution quality.
\begin{figure}[t]
    \centering
    \includegraphics[width=0.9\linewidth]{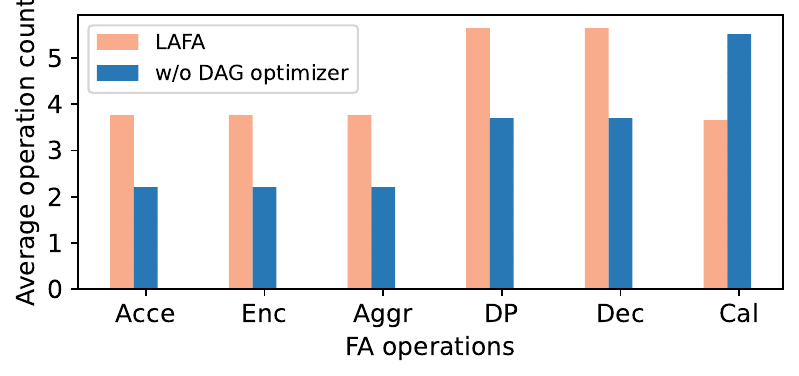}
    \caption{The impact of DAG optimizer in reducing operation count}
    \label{fig:stepCountAbl}
\end{figure}

The operation count comparison in Fig. \ref{fig:stepCountAbl} between the variant without the DAG optimizer and our full system reveals substantial efficiency gains:
Access to data, encryption, and aggregation operations per client are all reduced from 3.75 to 2.20, demonstrating the effectiveness of the DAG optimizer in reusing intermediate results across sub-queries.
The cost of ensuring DP and performing decryption is also lowered from 5.65 to 3.70, indicating that the DAG optimizer consolidates privacy operations rather than redundantly applying them to each isolated output. With all these resource-intensive operations being reduced, the lightweight calculation operation count increases from 3.65 to 5.50. This demonstrates that LAFA can offload complexity to lightweight post-processing to realize more computation-efficient execution. 


\section{Related work}

\subsection{LLM agents-assisted data analytics}
The emergence of LLMs has significantly advanced the landscape of automated data analytics.  
Recent developments have extended LLM usage from passive tools into active agent roles that can reason, plan, execute, and interact with data environments across multiple operations\cite{11183728}.
ReAct (Reason + Act) \cite{yao2023react} introduces a foundational prompting paradigm where LLMs interleave reasoning traces with executable actions, enabling them to solve complex tasks by interacting with external environments like Wikipedia or code sandboxes. 
It has inspired many subsequent agent-based designs in data analytics, especially those requiring iterative exploration and interaction. 

InsightPilot \cite{ding2023insightpilot} employs a tightly integrated LLM-agent and insight engine framework to automate exploratory data analytics (EDA). 
InfiAgent-DABench \cite{hu2024infiagent} further formalizes the evaluation of LLM agents in analytics tasks by introducing a benchmark and agent framework for data analytics from raw CSV files. 
AutoTQA \cite{zhu2024autotqa} expands the concept of LLM agents into a multi-agent framework for complex Tabular Question Answering (TQA) across multiple heterogeneous data sources. 
Other systems, such as \cite{kazemitabaar2024improving}, focus on improving steering and verification of LLM-driven data analytics through user-facing interfaces.

\subsection{Federated analytics}
FA has become a prominent approach for enabling privacy-preserving data analytics. 
Existing FA research addresses a range of query types, including those based on count-mean-sketch techniques \cite{Apple_DP_2017,roth-2019-honeycrisp}, frequency-related tasks \cite{Erlingsson_RAPPOR_2014,wang2022fedfpm,qin2016heavy,wang2024federated}, and graph metrics analytics and learning 
\cite{ye2020lf,liu2024edge,pan2022fedwalk,roth-2021-mycelium, pan2023lumos}. 

Honeycrisp \cite{roth-2019-honeycrisp} introduces a resilient committee-based framework for executing large-scale count-mean-sketch queries without relying on a trusted server. 
Building on this foundation, Orchard \cite{roth-2020-orchard} refines the approach by partitioning queries into distinct zones, each tailored to handle computations based on varying privacy requirements. 
Arboretum \cite{margolin-2023-arboretum} advances these concepts further by automatically optimizing query plans to efficiently support a wide array of analytics tasks and harnessing the computational resources of participant devices.

However, existing LLM systems lack privacy protection, assuming unrestricted data access. Existing FA systems only support specific, simple queries, which require data expertise to master, limiting their wide-scale deployment.
In contrast, LAFA offers privacy-preserving data analytics over decentralized data and complex queries in natural languages. 

\section{Conclusion}

In this work, we present LAFA, the first LLM-driven FA system that enables privacy-preserving, natural language-based data analytics over decentralized data sources. 
It bridges the gap between LLM-agent-based systems that lack privacy guarantees and FA frameworks that lack support for natural language input and exhibit deficiencies in handling complex queries. LAFA adopts a hierarchical multi-agent architecture that decomposes complex natural language queries into executable FA operations with high accuracy with high quality. To further optimize execution efficiency, it further introduces an optimizer agent to consolidate multiple raw DAGs into an optimized execution graph. 
Experimental results demonstrate that LAFA can generate executable FA operations with almost 100\% success rates while significantly lowering computation footprint. 

\bibliographystyle{IEEEtran}
\bibliography{ref.bib}
\section*{Appendix}
\subsection{Query generation}
\label{queryPrompt}

We adopt a retrieval-augmented generation (RAG) approach to encourage the generation of realistic and challenging queries for evaluation. We use the following prompt format:

``Based on the behaviors and use cases described in the documents/dataset, generate a list of realistic analytics queries that a privacy-preserving system might need to support. Each query should involve meaningful data analytics, can be complex that including multiple analytics intents, and may include comparisons across user groups, time periods, or feature categories. Avoid overly generic questions; queries should reflect plausible analytical tasks encountered in practice."

The RAG-based prompting framework enables us to construct 20 representative queries for each document/dataset that capture the richness and variety of federated analytics workloads while preserving alignment with practical utility and privacy constraints.

\subsection{Example queries}
\label{exampleQueries}

\textbf{AdultPii queries:}
\begin{enumerate}
\item What is the percentage of individuals with more than 12 years of education, and how does this group’s average salary compare to those with fewer years?
\item What is the average salary for each relationship category, and how does it change between individuals who are household heads and those who are not?
\end{enumerate}

\textbf{Apple queries:} 
\begin{enumerate}
\item What are the most frequently used emoji in messages this month, and how does their usage differ between users under 30 and over 50?
\item Which emoji are most common during working hours versus late at night, and what is the average number of emoji per message in each time period?
\end{enumerate}

\end{document}